\documentclass{article} 
\usepackage{iclr2026_conference,times}
\iclrfinalcopy 

\usepackage{amsmath,amsfonts,bm}

\def\eqref#1{equation~\ref{#1}}

\def\1{\bm{1}}

\DeclareMathAlphabet{\mathsfit}{\encodingdefault}{\sfdefault}{m}{sl}
\SetMathAlphabet{\mathsfit}{bold}{\encodingdefault}{\sfdefault}{bx}{n}

\usepackage{hyperref}
\usepackage{url}
\usepackage{lipsum}
\usepackage{wrapfig}

\usepackage{amsmath}      
\usepackage{amsfonts}
\usepackage{amssymb}      
\usepackage{mathdots}     
\usepackage{array}

\usepackage{graphicx}     
\usepackage{booktabs}     
\usepackage{multirow}     
\usepackage{adjustbox}    
\usepackage{subcaption}   

\usepackage{algpseudocode}  
\usepackage[ruled,vlined,linesnumbered]{algorithm2e} 

\usepackage{pifont}      
\usepackage{xspace}      

\usepackage{xcolor}      
\usepackage[skins]{tcolorbox} 
\usepackage{listings}

\usepackage{amsthm}

\newtheorem{theorem}{Theorem}[section]

\theoremstyle{definition}

\theoremstyle{remark}

\usepackage{enumitem}

\title{Unveiling the Potential of Diffusion Large Language Model in Controllable Generation}

\author{Zhen Xiong$^{1}$,\,\,\,Yujun Cai$^{2}$,\,\,\,Zhecheng Li$^{3}$,\,\,\,Yiwei Wang$^{4}$\\
$^{1}$University of Southern California \quad
$^{2}$University of Queensland \\
$^{3}$University of California, San Diego \quad
$^{4}$University of California, Merced \\
\href{https://eric2i.github.io/dLLM-CtrlGen}{\textcolor{magenta}{\texttt{eric2i.github.io/dLLM-CtrlGen}}}
}

\begin{document}

\maketitle

\begin{abstract}
Controllable generation is a fundamental task in NLP with many applications, providing a basis for function calling to agentic communication. However, even state-of-the-art autoregressive Large Language Models (LLMs) today exhibit unreliability when required to generate structured output. Inspired by the current new diffusion-based large language models (dLLM), we realize that the architectural difference, especially the global information-sharing mechanism for language modeling, may be the key to unlock next-level controllable generation. To explore the possibility, we propose \textbf{S}elf-adaptive \textbf{S}chema \textbf{S}caffolding ($S^3$), a novel framework that enables dLLM to stably generate reliable structured outputs (e.g., JSON) by utilizing its innate reverse reasoning capability and global context awareness. $S^3$ initiates a schematic template directly in the output context as a starting state for dLLM, offering a more robust and general method than intricate prompt optimization. Experiments demonstrate that our method substantially unlocks the dLLM’s potential in controllable generation in terms of structure adherence, content fidelity, and faithfulness. These results establish new perspectives and practical pathways for deploying language models in controllable generation tasks.
\end{abstract}

\section{Introduction}
Controllable generation is a fundamental task in the era of LLMs. It provides the foundation for stable tool use, agentic communication, and interaction with existing application programming interfaces (APIs). 
Existing works demonstrate that structured output still poses significant challenges even for state-of-the-art autoregressive LLMs. Many inspiring explorations have been conducted by previous researchers to address these challenges.

Prior autoregressive-based language model's methods pair a grammar-driven finite-state automata (FSA) with constrained decoding to enforce structural constraints during generation \citep{koo2024automata}. When no token satisfies the grammar, all beams are pruned and generation halts. More broadly, as instruction following has improved, practitioners have turned to prompt-engineering heuristics to elicit structurally compliant outputs. Yet hand-crafted prompts for diverse structural specifications are labor-intensive and yield inconsistent results across domains and complexity levels.

\begin{figure}[!bt]
    \centering
    \includegraphics[width=1.00\textwidth]{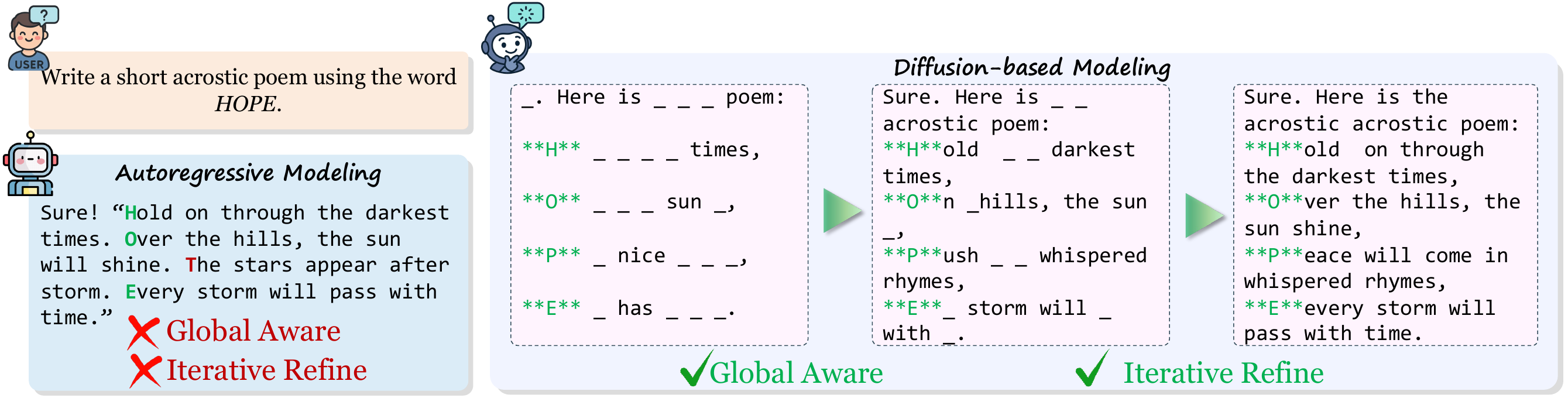}
    \caption{Illustrative comparison between autoregressive and diffusion-based language modeling on tasks requires specific global structure control and token-space planning in advance.}
    \label{fig:overview}
\end{figure}

Existing approaches share a fundamental limitation: they rely solely on language models' intrinsic capabilities without additional mechanisms to guide generation trajectories. This stems from architectural constraints of autoregressive models: (1) left-to-right generation prevents global structural coherence, as early tokens are generated without full sequence knowledge; (2) commitment to previously generated tokens limits backtracking when structural violations occur; (3) sequential dependencies inhibit simultaneous satisfaction of multiple constraints. Effective structured generation thus requires global sequence planning, iterative refinement, and parallel constraint satisfaction—capabilities that autoregressive architectures inherently lack.

Under this scenario, we realize that the diffusion-based large language model (dLLM) can serve as a natural alternative to traditional autoregressive (AR) generation by iteratively denoising corrupted inputs, enabling global context modeling potential and parallel token generation~\citep{arriola2025block,nie2025large,yu2025discrete}. Recent studies show that dLLMs can match AR models in instruction following, in-context learning, and math reasoning tasks~\citep{gong2024scaling,you2025llada}, while also offering enhanced controllability and faster inference~\citep{huang2025ctrldiff, labs2025mercury}. 

However, current open-sourced instruction-tuned dLLMs~\citep{nie2025large,zhu2025llada,dream2025} fail to produce well-structured outputs, often generating hallucinated content or breaking structural constraints. Additionally, their inference speed remains limited. Analysis reveals that existing implementations (e.g., semi-autoregressive approaches~\cite{nie2025large}) systematically undermine dLLMs' advantages in global awareness and parallel generation.

To unveil its full potential, we introduce a novel Self-adaptive Schema Scaffolding ($S^3$) method to fully unlock dLLM's potential in controllable generation. Specifically, $S^3$ inject a schematic template into the output context instead of instruction, providing a more \textit{compelling} language prior for dLLM during generation. With our method, dLLM can achieve significantly improved structured output quality. To quantify performance, we introduce a comprehensive framework that evaluates structured outputs along three key dimensions: structural adherence, content fidelity, and faithfulness (with detailed definitions provided in Section~\ref{sec:experiment}). Experimental results show that our method marginally improves the performance structured output using dLLM compared with the commonly used prompting strategy. 

Our main contributions can be summarized as follows: 1) We analyze the architectural advantages of diffusion-based large language models (dLLMs) compared with autoregressive models for controllable generation, focusing on prior's global attention mechanism and iterative refinement capabilities. 2) We propose Self-adaptive Schema Scaffolding ($S^3$), a training-free method that enables dLLMs to achieve higher structure output performance with fewer denoising steps. 3) We establish a comprehensive structure output evaluation framework focusing on structural adherence, content fidelity, and faithfulness metrics. Extensive experiments show that our $S^3$ achieves superior performance across all metrics with reduced computational complexity. Together, our work establishes a new perspective and practical solutions for deploying dLLMs for controllable generation tasks.

\section{Related Works}

\paragraph{Structured Output} requires models to generate in predefined formats (e.g., code, JSON, XML, or tables), supporting tasks such as entity extraction~\citep{li2024drs}, classification~\cite{huang2025appealcase}, and correlation prediction~\cite{xiong2025mapping}.
Grammar-constrained decoding enforces compliance with context-free grammars~\citep{geng2023grammar} or type systems~\citep{mundler2025type} by adjusting next-token probabilities, without task-specific fine-tuning. Alternatively, planning-based or two-stage strategies first predict intermediate structures, such as abstract syntax trees, and then realize the final output~\citep{wang2025slot}.
Most prior work relies on autoregressive LLMs for their strong language modeling and instruction-following ability~\citep{wei2022finetuned}. Diffusion-based models remain underexplored, motivating us to analyze their potential for structured generation and propose an evaluation framework covering structure compliance, content fidelity, and hallucination.

\paragraph{Autoregressive Large Language Models}
Autoregressive LLMs have become a general solution across NLP tasks, strengthened by advances such as longer context windows~\citep{liu2025comprehensive}, multimodal integration~\citep{han2025multimodal}, and test-time scaling~\citep{jaech2024openai,guo2025deepseek,gemini2025}. They achieve state-of-the-art results on benchmarks including MMLU~\citep{hendrycks2020measuring}, WebArena~\citep{zhou2023webarena}, and AIME~\citep{zhang2025aime}, but still face persistent challenges in hallucination and controllability.

\paragraph{Diffusion-based Language Models}
Diffusion models have recently been extended to discrete text, offering an alternative to autoregressive generation~\citep{li2023diffusion}. Research has introduced discrete score-based processes, refined noise schedules, and faster sampling methods, all aimed at improving efficiency and output quality. Building on these advances, large diffusion language models such as LLaDA~\citep{nie2025large,zhu2025llada} and Dream~\citep{dream2025} demonstrate strong instruction-following ability. Yet their reliance on multi-step denoising weakens the advantage of parallel generation and slows inference~\citep{israel2025accelerating}. Motivated by these limitations, we focus on structured output as a setting where diffusion models can better exploit their architecture, and propose an inference pipeline that accelerates generation for practical applications.

\section{Preliminary and Notations}

\subsection{Autoregressive Language Modeling}
Autoregressive large language models (LLMs) generate text by predicting the next token. During inference, instruction-tuned models~\citep{ouyang2022training, wei2022finetuned} produce a response $A$ to a query $Q$ by sampling from the conditional distribution
\begin{equation}
\log P_{\theta}(A|Q) = \sum_{t=1}^{|A|} \log P_{\theta}(a_t | a_{<t}, Q), \label{eq:ar}
\end{equation}
where $A = (a_1, \ldots, a_{|A|})$, $a_t$ is the token at position $t$, $a_{<t}$ are the preceding tokens, $|A|$ is the sequence length, and $\theta$ are model parameters.

Due to the sequential nature of autoregressive generation, there are two systematic limitations: \textit{models cannot directly access \textbf{future} token information during generation, and previously generated tokens cannot be \textbf{revised}.} These constraints may limit the model's performance on tasks that would benefit from lookahead planning or iterative refinement.

\subsection{Diffusion Language Modeling}
\label{sec:diffusion_language_modeling}

Diffusion models~\citep{ho2020denoising, nichol2021improved}, originally developed for continuous image generation tasks, have been adapted to discrete language modeling~\citep{austin2021structured, nie2025large, dream2025}. 

Diffusion-based language models consist of two key processes: a forward noising process and a reverse denoising process.
Given a tokenized sequence $\mathbf{x}_0$, the forward process progressively corrupts the sequence by masking tokens, producing increasingly noisy sequences $\mathbf{x}_t$ for $t\in[0,1]$ . As $t$ approaches $1$, more tokens are masked until $x_1$ becomes a fully masked sequence. The reverse process trains a neural network to predict the original tokens at masked position within $x_t$ for $t\in(0, 1]$. The pre-training objective can be formulated as:
\begin{equation}
    \min_{\phi} -\mathbb{E}_{t,x_0,x_t} \left[ \frac{1}{t} \sum_{i=1}^{L} \mathbf{1}[x_t^i = \mathbf{M}] \log P_{\phi}(x_0^i|x_t) \right] \label{eq:dllm-learning}
\end{equation}
where $\mathbf{M}$ denotes the mask token, $L$ is the sequence length, and $\phi$ represents the model parameters.

Recent research showed that diffusion large language model (dLLM) can be instruction-finetuned by concatenating ($\oplus$) user instruction $Q$ as fixed prefix to the masked target sequence $\mathbf{A}_t$ ($t=1$, all masked initially), enabling conditioned generation. During inference, a single denoising step that the dLLM take can be factorized as:
\begin{equation}
\log P_{\phi}(Q \oplus A) = \sum_{i=1}^{|A|} \mathbf{1}[a_i = \mathbf{M}]\log P_{\phi}(a_i | Q \oplus A_{t})  \label{eq:dllm-inference}
\end{equation}

Compared with AR models, the multi-step generation process can potentially be extended for iterative token edition and refinement~\citep{havasi2025edit}. More interestingly, the global attention mechanism of dLLM can largely improve its global context awareness and even including \textit{future} token planning. Therefore, in this paper, we exploit dLLM's superior global awareness and explore its potential for structured output.

\section{Methods}

In this section, we first establish a rigorous theoretical framework for structured output generation (Section~\ref{task_formulation}). 
Subsequently, we introduce our training-free pipeline, \textit{Schema Scaffolding} ($S^2$) (Section~\ref{schema_scaffolding}) and its improved version \textit{Self-adaptive Schema Scaffolding} ($S^3$)(Section~\ref{self_adaptive_schema_scaffolding}).

\subsection{Task Formalization}
\label{task_formulation}
Formally, we define structured output generation task as follows: given a user query $Q$, and a structural specification $S$, the objective is to generate a response $A$ that satisfies both the semantic requirements of $Q$ and the structural constraints defined by $S$. 
The structural specification $S$ can take various forms, including but not limited to: schema-based structures, format constraints, compositional structures, and domain-specific formats. 
The task can be mathematically formulated as a constrained optimization problem: $$A^* = \arg \max_{A \in \mathcal{A}(S)} P_{\text{LM}}(A | Q, S)$$
where $\mathcal{A}(S)$ represents the space of all valid outputs conforming to structure $S$ and $P_{\text{LM}}$ is a conditional langugage model.
In practice, searching entire $\mathcal{A}(S)$ is intractable due to the complexity of the token space.

\subsection{Schema Scaffolding}
\label{schema_scaffolding}

\begin{figure}[bt]
    \centering
    \includegraphics[width=\linewidth]{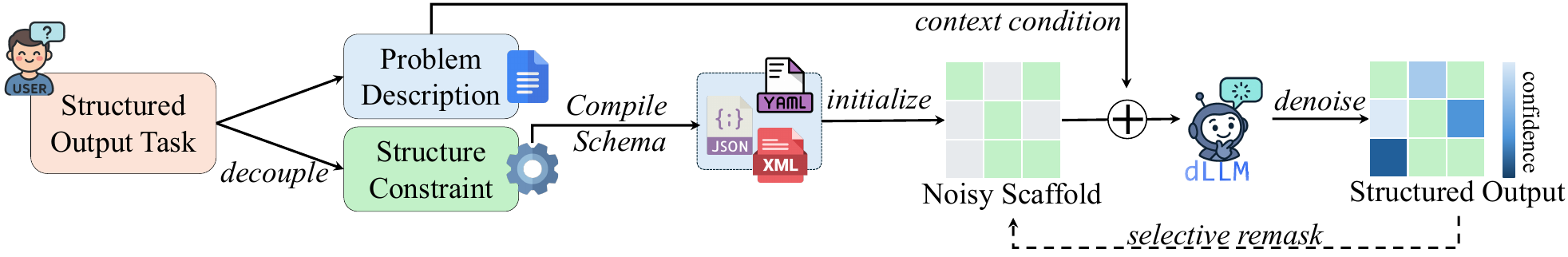}
    \caption{The overview of our method's pipeline. We begin by decomposing the original task instruction into two components: a problem description and a set of structural constraints. These constraints are compiled into a schema, which is then used to initialize a noisy scaffold where mask tokens serve as placeholders for missing content. The dLLM completes this scaffold by predicting the masked tokens, using the problem description as context to generate structured outputs. Additionally, we apply a selective remasking strategy that allows the model to iteratively refine its predictions and further improve generation quality.
}
    \label{fig:enter-label}
\end{figure}

Building on the theoretical insight we formulated in Section~\ref{sec:diffusion_language_modeling}, we propose \textit{schema scaffolding}, a training-free approach that explicitly incorporates structural constraints by pre-populating the generation context with structural templates.

The core idea of this approach is to transform unconstrained generation into a structured \textbf{fill-in-the-blank} task: In detail, our method operates by parsing the structural specification $S$ to identify invariant structural elements (e.g., brackets, delimiters, field names) and replacing variable content positions with mask tokens $\mathbf{M}$. 
This creates a structural scaffold $\textcolor{blue}{A_s}$ that constrains the model's generation space while preserving semantic flexibility. To formalize why this approach is effective for diffusion models, we establish the following result:

\begin{theorem}[Scaffold-Guided Denoising Convergence]
\label{thm:scaffold_convergence}
Let $\mathbf{x}_0$ be a target structured sequence and $\mathbf{x}_t$ be a partially masked sequence at timestep $t$ with scaffold $\mathcal{S}$ defining fixed structural positions. For a diffusion language model trained with objective (Eq.~\ref{eq:dllm-learning}), initializing the denoising process with structural scaffold $\mathcal{S}$ reduces the expected denoising error by:
$$\mathbb{E}[\|\hat{\mathbf{x}}_0 - \mathbf{x}_0\|_{\mathcal{M}}] \leq \mathbb{E}[\|\tilde{\mathbf{x}}_0 - \mathbf{x}_0\|_{\mathcal{M}}] \cdot \left(1 - \frac{|\mathcal{S}|}{L}\right)$$
where $\hat{\mathbf{x}}_0$ is generated with scaffolding, $\tilde{\mathbf{x}}_0$ is generated without scaffolding, $\|\cdot\|_{\mathcal{M}}$ denotes error over masked positions only, and $|\mathcal{S}|/L$ represents the scaffold coverage ratio.
\end{theorem}

The proof is deferred to Appendix~\ref{proof:scaffold_convergence}. This result confirms that scaffolding provides a principled way to guide the denoising process, with error reduction proportional to the scaffold coverage. It also confirms our later empirical findings where even minimal scaffolding achieves near-perfect structural adherence.

Formally, the structured generation objective of our proposed method here is:
\begin{align*}
A^* &= \arg \max_{A \in \mathcal{A}(S)} P_{\text{LM}}(A | Q, S) \\
&\approx \arg \max_{\textcolor{blue}{A_s} \in \mathcal{SC}} P_{\phi}(\textcolor{blue}{A_s} | Q) \\
&= \arg \max_{\textcolor{blue}{A_s} \in \mathcal{SC}} \sum_{a_i \in \textcolor{blue}{A_s}}\mathbf{1}[a_i = \mathbf{M}]\log P_{\phi}(a_i | Q, \textcolor{blue}{A_s})
\end{align*}

Where $\mathcal{SC} \subset \mathcal{A}(S)$ represents the constrained subspace of outputs sharing the structural template derived from $S$, and $\textcolor{blue}{A_s}$ denotes the scaffold with all non-masked tokens fixed except position $i$.

\subsection{Self-Adaptive Schema Scaffolding}
\label{self_adaptive_schema_scaffolding}

While our previous strategy constrains dLLM generation for improved structural compliance, it also introduces another challenge: \textit{how to determine the appropriate number of mask tokens for each variable content position?} 
Although we can build structured fill-in-the-blank templates in advance, predicting the required length for each variable field remains problematic without prior knowledge of the target content.

A straightforward solution is to allocate ample mask tokens for each variable position, expecting the model to use only what is needed. Yet our analysis (Sec.~\ref{dis:hallucination}) shows that dLLMs are sensitive to sequence length: longer scaffolds often distort generation quality instead of enabling selective usage, leading to under-utilization or hallucinated content. Another option is to introduce specialized padding tokens through fine-tuning. While appealing in principle, this violates our training-free objective and risks embedding dataset-specific biases that harm generalization to unseen domains.

Motivated by these intuitions, we propose an improved method that leverages the semantic token \texttt{null} as an placeholder. This approach preserves the training-free property while guiding dLLMs to naturally represent absent or variable-length content with \texttt{null} tokens.

Formally, we extend our scaffolding framework to incorporate adaptive length management:
\begin{align*}
    A^*\approx \arg \max_{\textcolor{blue}{A_s} \in \mathcal{SC}} \sum_{a_i \in \textcolor{blue}{A_s}}\mathbf{1}[a_i = \mathbf{M}]\log P_{\phi}(a_i | \textcolor{red}{Q^+}, \textcolor{blue}{A_s})
\end{align*}
where $\textcolor{red}{Q^+}$ represents our augmented prompt which guide the model to adopt $\texttt{null}$ tokens to indicate absent values.

Our approach here empirically transforms the fixed-length scaffolding problem into an adaptive generation task where dLLMs can naturally handle variable-length fields and missing values. Further experimental results demonstrate that Self-Adaptive Schema Scaffolding significantly improves overall structured output quality, particularly in scenarios involving optional fields or variable-length content.

\section{Experiments}
\label{sec:experiment}

\subsection{Implementation Details}

We use the LLaDA~\citep{nie2025large} model as our primary dLLM for experiments and the WikiBio dataset~\citep{lebret2016wikibio} as our dataset. For reproducibility, we set the decoding temperature to zero during inference. See Appendix~\ref{appendix:impl} for more details.

\subsection{Evaluation Framework}
\label{sec:evaluation_framework}

Given the unique characteristics of structured output generation, we argue that traditional accuracy-based metrics alone are inadequate for capturing the full spectrum of model performance. Thus, we propose a comprehensive evaluation framework that assesses outputs across three key dimensions: \textbf{Structural Adherence}, \textbf{Content Fidelity}, and \textbf{Faithfulness}. This multi-dimensional benchmark enables a more nuanced and reliable assessment of a model's ability to generate coherent, informative, and trustworthy structured texts.

\textbf{Structural Adherence} measures how well-generated outputs conform to the target schema. Specifically, we define:

\vspace{-2pt}
\begin{itemize}[leftmargin=25pt, topsep=0.5pt, itemsep=0.5pt,]
    \setlength{\itemsep}{2pt}  
    \item \textit{Structure Validity (SV)}: the proportion of outputs that are syntactically valid and parse without errors, capturing basic structural correctness.
    \item \textit{Field Completeness (FC)}: the percentage of required fields that are correctly populated, indicating whether the model includes all mandatory schema components.
    \item \textit{Schema Compliance (SC)}: the strictest structural metric, measuring the proportion of outputs that fully adhere to the predefined schema, including correct data types, value constraints, and nested structures.
\end{itemize}
\vspace{-2pt}

\textbf{Content Fidelity} evaluates the semantic accuracy of information within structurally valid outputs. We consider:

\vspace{-2pt}
\begin{itemize}[leftmargin=25pt, topsep=0.5pt, itemsep=0.5pt,]
    \item \textit{Precision/Recall (PR/RE)}: precision and recall scores computed over individual field types, providing a detailed view of model behavior across content categories.

    \item \textit{F1 Score (F1)}: the harmonic mean of field-level precision and recall, computed using both exact match and fuzzy match strategies to accommodate minor textual variations that preserve semantic meaning.
\end{itemize}
\vspace{-2pt}

\textbf{Faithfulness} assesses the degree to which generated content remains grounded in the source input, which is especially critical in extraction settings.

\vspace{-2pt}
\begin{itemize}[leftmargin=25pt, topsep=0.5pt, itemsep=0.5pt,]
    \item \textit{Hallucination Rate (HR)}: the proportion of output fields that include information not present in or not reasonably inferable from the source text, directly reflecting the model’s factual consistency.
\end{itemize}
\vspace{-2pt}

Together, these metrics support systematic comparisons across models and highlight specific areas for targeted improvement in structured output generation.

\subsection{Main Results}
This section presents a comprehensive evaluation of our approaches against baseline methods across three critical dimensions: structural adherence, content fidelity, and faithfulness.

\subsubsection{Structural Adherence}
\begin{figure}[!bt]
\centering
\includegraphics[width=\textwidth]{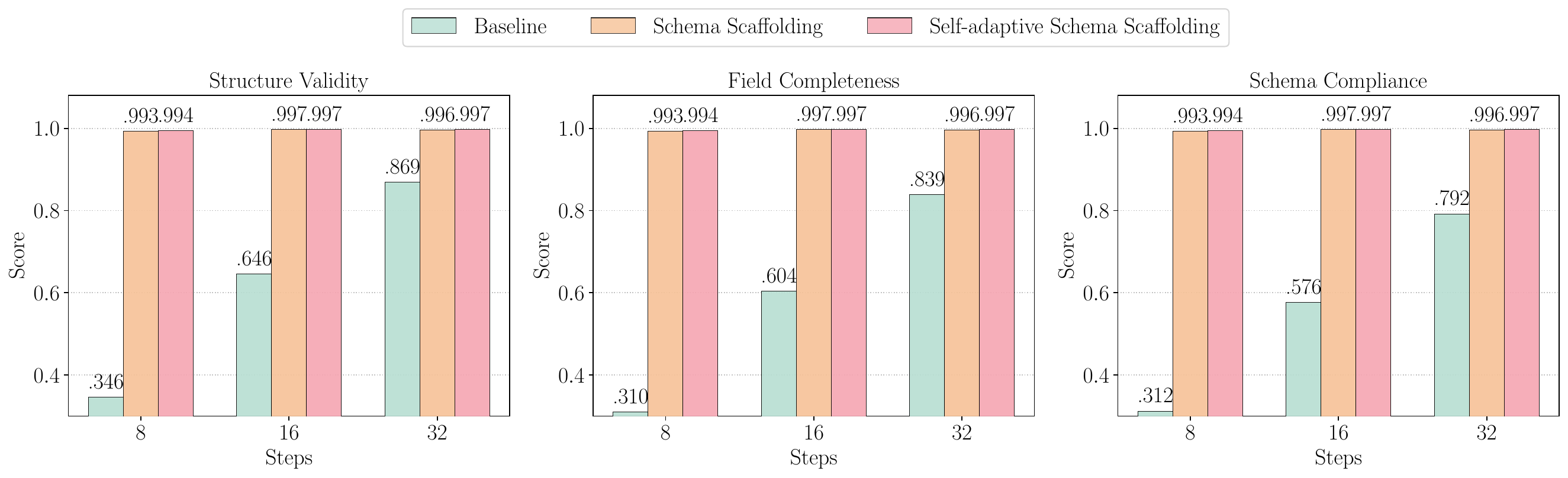}
\caption{\textbf{Structural adherence} comparison across denoising steps and methods. Results show consistent improvements across all metrics using our schema scaffolding approaches, with near-perfect performance achieved in fewer steps.}
\label{fig:structural_adherence}
\end{figure}

Direct prompting dLLM is inadequate for structural constraint. Our baseline evaluation reveals performance consistently below 65\% across all structural metrics—\textit{Schema Validity}, \textit{Field Completeness}, and \textit{Schema Compliance}. Even with 32 denoising steps, the maximal score among these three critical metrics is above 87\%, far lower than the expectation for realistic utility.(Fig.~\ref{fig:structural_adherence}).

Our schema scaffolding methods marginally outperform the baseline method in terms of structural adherence. Both vanilla and self-adaptive variants enable near-perfect structural adherence with as few as 8 denoising steps, achieving performance saturation at 16 steps. This dramatic improvement carries practical significance beyond accuracy: since diffusion model inference scales linearly with denoising steps, our approach simultaneously reduces generation latency while enhancing structural quality.
These findings establish that schema scaffolding addresses the fundamental barrier to deploying diffusion-based language models for structured output generation, elevating them from impractical to highly effective for real-world applications.

\subsubsection{Content Fidelity}

\begin{figure}[!bt]
    \centering
    \includegraphics[width=\textwidth]{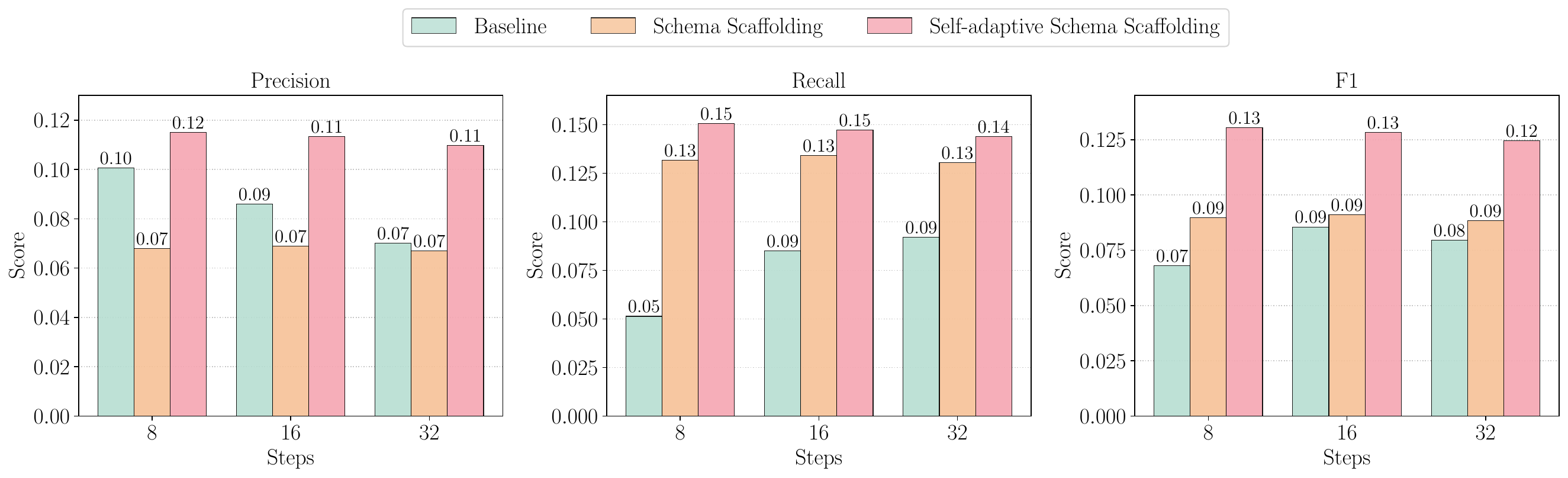}
    \caption{\textbf{Content fidelity} comparison across denoising steps and methods. Our self-adaptive schema scaffolding consistently achieves the highest precision, recall, and F1 score across all settings.}
    \label{fig:content_fidelity}
\end{figure}

For content fidelity, additional denoising steps counterintuitively do not guarantee improved content accuracy and we observe that performance sometimes degrades with extended iteration (Fig.~\ref{fig:content_fidelity}). This pattern reflects the diffusion model may deviation from optimal solutions during extended iterative reverse process, what we term the "overthinking" phenomenon.

\begin{wraptable}{r}{0.50\textwidth}
  \centering
  \begingroup
    \small 
    \setlength{\tabcolsep}{13pt}     
    \renewcommand{\arraystretch}{1.3} 
    \begin{tabular}{@{}cccc@{}}
      \toprule
      \multirow{2}{*} {Method} & \multicolumn{3}{c} {Computation Budget} \\ 
      \cmidrule{2-4} & 8 Steps & 16 Steps & 32 Steps \\
      \midrule
      Baseline & $0.404$ & $0.403$ & $0.409$ \\
      $S^2$    & $0.465$ & $0.463$ & $0.463$ \\
      $S^3$    & $\textbf{0.340}$ & $\textbf{0.331}$ & $\textbf{0.331}$ \\
      \bottomrule
    \end{tabular}
  \endgroup
\caption{Hallucination rate  (\textit{lower is better}) comparison across various denoising steps. Our self‑adaptive schema scaffolding ($S^3$) method consistently achieves the lowest hallucination rates, indicating superior faithfulness. Boldface indicates the best results.}
\label{tab:hallucination-rate}
\end{wraptable}

Vanilla Schema Scaffolding demonstrates clear improvements in recall and F1 score relative to baseline, indicating enhanced coverage of relevant content. However, precision suffers notably across all denoising configurations. This trade-off emerges from the model's compensatory behavior: when constrained by rigid schema requirements, it over-generates tokens to fill all reserved slots, particularly problematic when content length varies significantly across examples.

Our Self-adaptive Schema Scaffolding ($S^3$) resolves this tricky situation. By incorporating adaptive \texttt{null} token for surplus slots, the method prevents over-generation while maintaining comprehensive coverage. This simple yet effective recipe yields substantial improvements across all three metrics, establishing robust content fidelity without sacrificing structural compliance.

\subsubsection{Faithfulness}

Our self-adaptive schema scaffolding ($S^3$) method demonstrates superior faithfulness performance, consistently achieving the lowest hallucination rates across all denoising steps (Tab.~\ref{tab:hallucination-rate}). This clear advantage over baseline approaches establishes the superiority of our method in maintaining factual grounding while generating structured outputs.

Interestingly, vanilla schema scaffolding reveals elevated hallucination rates compared to baseline methods. The worse performance in faithfulness represents a critical weakness that challenges our original design of the Schema Scaffolding approach. Our analysis reveals that this hallucination issue stems from the distributional shifts introduced by the imposed structural constraints, which create a token distribution that diverges from the model's pre-training reverse process. Specifically, the rigid schema acts as a structural prior that misaligned with the diffusion language model's learned denoising process. This distributional mismatch forces the model into suboptimal denoising trajectories. When the structural prior demands content generation beyond what can be grounded in the source text, the model defaults to fabricating plausible tokens to satisfy schema requirements—a behavior that contradicts its training objective of faithful reconstruction. 

Our self-adaptive approach mitigates this fundamental conflict by allowing the model to acknowledge missing information through \texttt{null} tokens, thus maintaining alignment with its pre-trained denoising capabilities while respecting structural constraints.

\subsubsection{Ablation Study}

To demonstrate the effectiveness of our approach, we conduct ablation studies comparing our method against baseline models that are incrementally enhanced with different techniques (Tab.\ref{tab:ablation_metrics}). While few-shot learning and template-as-guidance approaches improve dLLM's structural adherence and faithfulness, their performance on fidelity remains inconsistent and shows limited improvement. In contrast, our zero-shot method achieves superior data efficiency and computational efficiency. Despite requiring fewer denoising steps, $S^3$ delivers remarkably stable structural adherence, along with marginal improvements in both fidelity and faithfulness.

\begin{table}[!ht]
\centering
\begingroup
 \small 
    \setlength{\tabcolsep}{12.25pt}     
    \renewcommand{\arraystretch}{1.225} 
    \begin{tabular}{@{}lccccccc@{}}
    
    \toprule
    \multirow{2}{*}{\textbf{Experiment}} & \multicolumn{3}{c}{\textbf{Structural Adherence}} & \multicolumn{3}{c}{\textbf{Content Fidelity}} & \textbf{Faithfulness\,} \\
    \cmidrule{2-8} & SV$\uparrow$ & FC$\uparrow$ & SC$\uparrow$ & PR$\uparrow$ & RE$\uparrow$ & F1$\uparrow$ & HR$\downarrow$ \\
    \midrule
    
    \multicolumn{8}{c}{8 Steps} \\
    \midrule
    baseline & 0.346 & 0.310 & 0.312 & 0.101 & 0.051 & 0.068 & 0.404 \\
    \quad w/ few-shots & 0.471 & 0.447 & 0.443 & 0.086 & 0.056 & 0.068 & 0.366 \\
    \quad w/ template & 0.475 & 0.432 & 0.431 & 0.088 & 0.081 & 0.084 & 0.388 \\
    $S^3$ (ours) & \textbf{0.994} & \textbf{0.994} & \textbf{0.994} & \textbf{0.115} & \textbf{0.151} & \textbf{0.130} & \textbf{0.340} \\
    \midrule
    
    \multicolumn{8}{c}{16 Steps} \\
    \midrule
    baseline & 0.646 & 0.604 & 0.576 & 0.086 & 0.085 & 0.086 & 0.403 \\
    \quad w/ few-shots & 0.735 & 0.713 & 0.674 & 0.084 & 0.087 & 0.085 & 0.371 \\
    \quad w/ template & 0.794 & 0.734 & 0.738 & 0.091 & 0.139 & 0.110 & 0.390 \\
    $S^3$ (ours) & \textbf{0.997} & \textbf{0.997} & \textbf{0.997} & \textbf{0.113} & \textbf{0.147} & \textbf{0.128} & \textbf{0.331} \\
    \midrule
    
    \multicolumn{8}{c}{32 Steps} \\
    \midrule
    baseline & 0.869 & 0.839 & 0.792 & 0.070 & 0.092 & 0.080 & 0.409 \\
    \quad w/ few-shots & 0.890 & 0.870 & 0.824 & 0.092 & 0.114 & 0.101 & 0.358 \\
    \quad w/ template & 0.909 & 0.886 & 0.882 & 0.091 & \textbf{0.160} & 0.116 & 0.384 \\
    $S^3$ (ours) & \textbf{0.997} & \textbf{0.997} & \textbf{0.997} & \textbf{0.110} & 0.144 & \textbf{0.125} & \textbf{0.331} \\
    \bottomrule
    \end{tabular}%
\endgroup
\caption{Ablation study results comparing different experimental configurations. Boldface indicates the best results. We use 3 examples for few-shot learning (\cite{wei2022finetuned}). To follow previous practice, we also extend the instruction with a complete schema (\cite{wang2025slot}) as guidance. Our method consistently outperforms these alternative techniques, indicating a nontrivial and superior improvement.}
\label{tab:ablation_metrics}
\end{table}

\section{Discussion}
\subsection{Robustness against Hallucination}
\label{dis:hallucination}
Similar to autoregressive language models, diffusion-based language models (dLLMs) also suffer from hallucination—generating nonfactual content, faulty reasoning, or unsupported conclusions. Unlike open-ended dialogue or creative text generation, hallucination in structured output is particularly detrimental, as it directly undermines the reliability and trustworthiness of the output—qualities that are central to these tasks.

Our vanilla schema scaffolding method enforces a strong structural constraint, which inevitably interferes with the natural generation trajectory of dLLMs. To mitigate the resulting hallucination, we observe that it is possible to transform unfamiliar test-time scenarios into familiar training-time cases by providing sufficient in-context guidance. Thus, we introduce the notion of a special token, \texttt{null}, as a flexible placeholder. Once the model adapts this convention, it can fill otherwise empty slots without resorting to fabricated content, despite not being explicitly trained to use padding tokens. This simple yet effective prior guidance enables dLLMs to substantially reduce hallucination, and it inspired the formulation of our improved $S^3$ method.

\subsection{Complexity Analysis}
\label{dis:complexity}
An existing bottleneck of dLLMs lies in their inference speed. Empirically, the multi-step denoising process introduces a latency that grows positively with the number of diffusion steps. Within each step, the global attention computation incurs a quadratic cost with respect to the total context length $L$. As a result, the overall computational complexity of the reverse process scales as $\mathcal{O}(L^3)$.
To mitigate this cost, some implementations adopt a semi-autoregressive decoding scheme with block-wise KV-caching, which partially reduces the computational burden. However, this design still compromises the core parallelism advantages of diffusion-based decoding. 

For structured generation tasks, where the output structure is known or can be approximated, our proposed method $S^3$ introduces an alternative perspective and initialize the reverse process from a partially denoised state rather than a fully random one. This \textit{warm-start} initialization serves as a structural prior, effectively providing a language scaffold that accelerates generation and enhances controllability. Asymptotically, $S^3$ reduces the decoding complexity to $\mathcal{O}(nL^2)$, where $n$ is a tunable hyperparameter that remains significantly smaller than $L$ in practice.

\section{Conclusion}

In this paper, we explore the potential of diffusion large language models' global awareness for controllable generation of structured output. We propose the novel Self-adaptive Schema Scaffolding method ($S^3$) that guides dLLMs to adaptively generate fully controllable structured output by manipulating the reverse process and leveraging the innate global attention mechanism.
Our comprehensive evaluation framework demonstrates that $S^3$ achieves superior structural adherence, content fidelity, and reduced hallucination rates. Through complexity analysis, we show that our approach maintains computational efficiency while enabling higher levels of controllability and hallucination control.
We believe these findings establish dLLMs as a promising alternative for controllable generation tasks.

\bibliography{iclr2026_conference}

\begin{thebibliography}{33}
\providecommand{\natexlab}[1]{#1}
\providecommand{\url}[1]{\texttt{#1}}
\expandafter\ifx\csname urlstyle\endcsname\relax
  \providecommand{\doi}[1]{doi: #1}\else
  \providecommand{\doi}{doi: \begingroup \urlstyle{rm}\Url}\fi

\bibitem[Arriola et~al.(2025)Arriola, Gokaslan, Chiu, Yang, Qi, Han, Sahoo, and Kuleshov]{arriola2025block}
Marianne Arriola, Aaron Gokaslan, Justin~T Chiu, Zhihan Yang, Zhixuan Qi, Jiaqi Han, Subham~Sekhar Sahoo, and Volodymyr Kuleshov.
\newblock Block diffusion: Interpolating between autoregressive and diffusion language models.
\newblock \emph{arXiv preprint arXiv:2503.09573}, 2025.

\bibitem[Austin et~al.(2021)Austin, Johnson, Ho, Tarlow, and Van Den~Berg]{austin2021structured}
Jacob Austin, Daniel~D Johnson, Jonathan Ho, Daniel Tarlow, and Rianne Van Den~Berg.
\newblock Structured denoising diffusion models in discrete state-spaces.
\newblock \emph{Advances in neural information processing systems}, 34:\penalty0 17981--17993, 2021.

\bibitem[{Gemini Team, Google}(2025)]{gemini2025}
{Gemini Team, Google}.
\newblock Gemini 2.5: Pushing the frontier with advanced reasoning, multimodality, long context, and next generation agentic capabilities.
\newblock Technical report, Google DeepMind, June 2025.
\newblock URL \url{https://storage.googleapis.com/deepmind-media/gemini/gemini_v2_5_report.pdf}.
\newblock Technical Report.

\bibitem[Geng et~al.(2023)Geng, Josifoski, Peyrard, and West]{geng2023grammar}
Saibo Geng, Martin Josifoski, Maxime Peyrard, and Robert West.
\newblock Grammar-constrained decoding for structured nlp tasks without finetuning.
\newblock \emph{arXiv preprint arXiv:2305.13971}, 2023.

\bibitem[Gong et~al.(2024)Gong, Agarwal, Zhang, Ye, Zheng, Li, An, Zhao, Bi, Han, et~al.]{gong2024scaling}
Shansan Gong, Shivam Agarwal, Yizhe Zhang, Jiacheng Ye, Lin Zheng, Mukai Li, Chenxin An, Peilin Zhao, Wei Bi, Jiawei Han, et~al.
\newblock Scaling diffusion language models via adaptation from autoregressive models.
\newblock \emph{arXiv preprint arXiv:2410.17891}, 2024.

\bibitem[Guo et~al.(2025)Guo, Yang, Zhang, Song, Zhang, Xu, Zhu, Ma, Wang, Bi, et~al.]{guo2025deepseek}
Daya Guo, Dejian Yang, Haowei Zhang, Junxiao Song, Ruoyu Zhang, Runxin Xu, Qihao Zhu, Shirong Ma, Peiyi Wang, Xiao Bi, et~al.
\newblock Deepseek-r1: Incentivizing reasoning capability in llms via reinforcement learning.
\newblock \emph{arXiv preprint arXiv:2501.12948}, 2025.

\bibitem[Han et~al.(2025)Han, Mubarak, Baimagambetov, Polatidis, and Baker]{han2025multimodal}
Longzhen Han, Awes Mubarak, Almas Baimagambetov, Nikolaos Polatidis, and Thar Baker.
\newblock Multimodal large language models: A survey.
\newblock \emph{arXiv preprint arXiv:2506.10016}, 2025.

\bibitem[Havasi et~al.(2025)Havasi, Karrer, Gat, and Chen]{havasi2025edit}
Marton Havasi, Brian Karrer, Itai Gat, and Ricky~TQ Chen.
\newblock Edit flows: Flow matching with edit operations.
\newblock \emph{arXiv preprint arXiv:2506.09018}, 2025.

\bibitem[Hendrycks et~al.(2020)Hendrycks, Burns, Basart, Zou, Mazeika, Song, and Steinhardt]{hendrycks2020measuring}
Dan Hendrycks, Collin Burns, Steven Basart, Andy Zou, Mantas Mazeika, Dawn Song, and Jacob Steinhardt.
\newblock Measuring massive multitask language understanding.
\newblock \emph{arXiv preprint arXiv:2009.03300}, 2020.

\bibitem[Ho et~al.(2020)Ho, Jain, and Abbeel]{ho2020denoising}
Jonathan Ho, Ajay Jain, and Pieter Abbeel.
\newblock Denoising diffusion probabilistic models.
\newblock \emph{Advances in Neural Information Processing Systems}, 33:\penalty0 6840--6851, 2020.

\bibitem[Huang \& Tang(2025)Huang and Tang]{huang2025ctrldiff}
Chihan Huang and Hao Tang.
\newblock Ctrldiff: Boosting large diffusion language models with dynamic block prediction and controllable generation.
\newblock \emph{arXiv preprint arXiv:2505.14455}, 2025.

\bibitem[Huang et~al.(2025)Huang, Guo, Wu, Li, Liu, Yin, Sun, Wu, and Kuang]{huang2025appealcase}
Yuting Huang, Meitong Guo, Yiquan Wu, Ang Li, Xiaozhong Liu, Keting Yin, Changlong Sun, Fei Wu, and Kun Kuang.
\newblock Appealcase: A dataset and benchmark for civil case appeal scenarios.
\newblock \emph{arXiv preprint arXiv:2505.16514}, 2025.

\bibitem[Israel et~al.(2025)Israel, Broeck, and Grover]{israel2025accelerating}
Daniel Israel, Guy Van~den Broeck, and Aditya Grover.
\newblock Accelerating diffusion llms via adaptive parallel decoding.
\newblock \emph{arXiv preprint arXiv:2506.00413}, 2025.

\bibitem[Jaech et~al.(2024)Jaech, Kalai, Lerer, Richardson, El-Kishky, Low, Helyar, Madry, Beutel, Carney, et~al.]{jaech2024openai}
Aaron Jaech, Adam Kalai, Adam Lerer, Adam Richardson, Ahmed El-Kishky, Aiden Low, Alec Helyar, Aleksander Madry, Alex Beutel, Alex Carney, et~al.
\newblock Openai o1 system card.
\newblock \emph{arXiv preprint arXiv:2412.16720}, 2024.

\bibitem[Koo et~al.(2024)Koo, Liu, and He]{koo2024automata}
Terry Koo, Frederick Liu, and Luheng He.
\newblock Automata-based constraints for language model decoding.
\newblock \emph{arXiv preprint arXiv:2407.08103}, 2024.

\bibitem[Labs et~al.(2025)Labs, Khanna, Kharbanda, Li, Varma, Wang, Birnbaum, Luo, Miraoui, Palrecha, et~al.]{labs2025mercury}
Inception Labs, Samar Khanna, Siddhant Kharbanda, Shufan Li, Harshit Varma, Eric Wang, Sawyer Birnbaum, Ziyang Luo, Yanis Miraoui, Akash Palrecha, et~al.
\newblock Mercury: Ultra-fast language models based on diffusion.
\newblock \emph{arXiv preprint arXiv:2506.17298}, 2025.

\bibitem[Lebret et~al.(2016)Lebret, Grangier, and Auli]{lebret2016wikibio}
R\'emi Lebret, David Grangier, and Michael Auli.
\newblock Generating text from structured data with application to the biography domain.
\newblock \emph{CoRR}, abs/1603.07771, 2016.
\newblock URL \url{http://arxiv.org/abs/1603.07771}.

\bibitem[Li et~al.(2023)Li, Zhou, Zhao, and Wen]{li2023diffusion}
Yifan Li, Kun Zhou, Wayne~Xin Zhao, and Ji-Rong Wen.
\newblock Diffusion models for non-autoregressive text generation: A survey.
\newblock \emph{arXiv preprint arXiv:2303.06574}, 2023.

\bibitem[Li et~al.(2024)Li, Wang, Hooi, Cai, Peng, and Chang]{li2024drs}
Zhecheng Li, Yiwei Wang, Bryan Hooi, Yujun Cai, Nanyun Peng, and Kai-Wei Chang.
\newblock Drs: Deep question reformulation with structured output.
\newblock \emph{arXiv preprint arXiv:2411.17993}, 2024.

\bibitem[Liu et~al.(2025)Liu, Zhu, Bai, He, Liao, Que, Wang, Zhang, Zhang, Zhang, et~al.]{liu2025comprehensive}
Jiaheng Liu, Dawei Zhu, Zhiqi Bai, Yancheng He, Huanxuan Liao, Haoran Que, Zekun Wang, Chenchen Zhang, Ge~Zhang, Jiebin Zhang, et~al.
\newblock A comprehensive survey on long context language modeling.
\newblock \emph{arXiv preprint arXiv:2503.17407}, 2025.

\bibitem[M{\"u}ndler et~al.(2025)M{\"u}ndler, He, Wang, Sen, Song, and Vechev]{mundler2025type}
Niels M{\"u}ndler, Jingxuan He, Hao Wang, Koushik Sen, Dawn Song, and Martin Vechev.
\newblock Type-constrained code generation with language models.
\newblock \emph{Proceedings of the ACM on Programming Languages}, 9\penalty0 (PLDI):\penalty0 601--626, 2025.

\bibitem[Nichol \& Dhariwal(2021)Nichol and Dhariwal]{nichol2021improved}
Alex Nichol and Prafulla Dhariwal.
\newblock Improved denoising diffusion probabilistic models.
\newblock \emph{International Conference on Machine Learning}, 139:\penalty0 8162--8171, 2021.

\bibitem[Nie et~al.(2025)Nie, Zhu, You, Zhang, Ou, Hu, Zhou, Lin, Wen, and Li]{nie2025large}
Shen Nie, Fengqi Zhu, Zebin You, Xiaolu Zhang, Jingyang Ou, Jun Hu, Jun Zhou, Yankai Lin, Ji-Rong Wen, and Chongxuan Li.
\newblock Large language diffusion models.
\newblock \emph{arXiv preprint arXiv:2502.09992}, 2025.

\bibitem[Ouyang et~al.(2022)Ouyang, Wu, Jiang, Almeida, Wainwright, Mishkin, Zhang, Agarwal, Slama, Ray, et~al.]{ouyang2022training}
Long Ouyang, Jeff Wu, Xu~Jiang, Diogo Almeida, Carroll~L Wainwright, Pamela Mishkin, Chong Zhang, Sandhini Agarwal, Katarina Slama, Alex Ray, et~al.
\newblock Training language models to follow instructions with human feedback.
\newblock \emph{Advances in Neural Information Processing Systems}, 35:\penalty0 27735--27756, 2022.

\bibitem[Wang et~al.(2025)Wang, Shen, Mishra, Xu, Teng, and Ding]{wang2025slot}
Darren Yow-Bang Wang, Zhengyuan Shen, Soumya~Smruti Mishra, Zhichao Xu, Yifei Teng, and Haibo Ding.
\newblock Slot: Structuring the output of large language models.
\newblock \emph{arXiv preprint arXiv:2505.04016}, 2025.

\bibitem[Wei et~al.(2022)Wei, Bosma, Zhao, Guu, Yu, Lester, Du, Dai, and Le]{wei2022finetuned}
Jason Wei, Maarten Bosma, Vincent~Y Zhao, Kelvin Guu, Adams~Wei Yu, Brian Lester, Nan Du, Andrew~M Dai, and Quoc~V Le.
\newblock Finetuned language models are zero-shot learners.
\newblock In \emph{International Conference on Learning Representations (ICLR)}, 2022.

\bibitem[Xiong et~al.(2025)Xiong, Cai, Li, and Wang]{xiong2025mapping}
Zhen Xiong, Yujun Cai, Zhecheng Li, and Yiwei Wang.
\newblock Mapping the minds of llms: A graph-based analysis of reasoning llm.
\newblock \emph{arXiv preprint arXiv:2505.13890}, 2025.

\bibitem[Ye et~al.(2025)Ye, Xie, Zheng, Gao, Wu, Jiang, Li, and Kong]{dream2025}
Jiacheng Ye, Zhihui Xie, Lin Zheng, Jiahui Gao, Zirui Wu, Xin Jiang, Zhenguo Li, and Lingpeng Kong.
\newblock Dream 7b, 2025.
\newblock URL \url{https://hkunlp.github.io/blog/2025/dream}.

\bibitem[You et~al.(2025)You, Nie, Zhang, Hu, Zhou, Lu, Wen, and Li]{you2025llada}
Zebin You, Shen Nie, Xiaolu Zhang, Jun Hu, Jun Zhou, Zhiwu Lu, Ji-Rong Wen, and Chongxuan Li.
\newblock Llada-v: Large language diffusion models with visual instruction tuning.
\newblock \emph{arXiv preprint arXiv:2505.16933}, 2025.

\bibitem[Yu et~al.(2025)Yu, Li, and Wang]{yu2025discrete}
Runpeng Yu, Qi~Li, and Xinchao Wang.
\newblock Discrete diffusion in large language and multimodal models: A survey.
\newblock \emph{arXiv preprint arXiv:2506.13759}, 2025.

\bibitem[Zhang(2025)]{zhang2025aime}
Di~Zhang.
\newblock Aime\_1983\_2024 (revision 6283828), 2025.
\newblock URL \url{https://huggingface.co/datasets/di-zhang-fdu/AIME_1983_2024}.

\bibitem[Zhou et~al.(2023)Zhou, Xu, Zhu, Zhou, Lo, Sridhar, Cheng, Ou, Bisk, Fried, et~al.]{zhou2023webarena}
Shuyan Zhou, Frank~F Xu, Hao Zhu, Xuhui Zhou, Robert Lo, Abishek Sridhar, Xianyi Cheng, Tianyue Ou, Yonatan Bisk, Daniel Fried, et~al.
\newblock Webarena: A realistic web environment for building autonomous agents.
\newblock \emph{arXiv preprint arXiv:2307.13854}, 2023.

\bibitem[Zhu et~al.(2025)Zhu, Wang, Nie, Zhang, Wu, Hu, Zhou, Chen, Lin, Wen, et~al.]{zhu2025llada}
Fengqi Zhu, Rongzhen Wang, Shen Nie, Xiaolu Zhang, Chunwei Wu, Jun Hu, Jun Zhou, Jianfei Chen, Yankai Lin, Ji-Rong Wen, et~al.
\newblock Llada 1.5: Variance-reduced preference optimization for large language diffusion models.
\newblock \emph{arXiv preprint arXiv:2505.19223}, 2025.

\end{thebibliography}
\bibliographystyle{iclr2026_conference}

\newpage
\appendix

\section{Implementation Details}
\label{appendix:impl}
We use the \texttt{GSAI-ML/LLaDA-1.5} diffusion large language model, loaded from HuggingFace official with \texttt{bfloat16} for optimal memory efficiency and performance. All experiments were conducted on one single NVIDIA RTX 4090 GPU. We set \texttt{max\textunderscore new\textunderscore tokens=128}, which suffices all structured output task, and \texttt{temperature=0} for reproducibility. The WikiBio dataset serves as the primary benchmark dataset. 

\section{Remasking Strategy}
\label{appendix:remask}

By default, dLLMs can generate all tokens within the full context in parallel through a single inference step. However, for many general NLP tasks, this strategy is often suboptimal. For example, simultaneously generating all tokens that form a complete solution to a math problem is significantly harder than doing so incrementally via the autoregressive approach. To address this issue, prior works have explored various remasking strategies to bridge the gap.

One common strategy is the block-wise masking, which employs a sliding-window mechanism. The context is divided into blocks, and at each iteration, only the tokens within the current window are taken into consideration, while those in following blocks are remasked for future regeneration. Thus, this strategy is also considered semi-autoregressive. However, empirical results show that using a block size of one, which essentially reverting to a fully autoregressive process, often yields the best performance. We consider such a strategy largely eliminates the parallelism advantage of diffusion-based models.

Another type of works focus on low-confidence remasking, which selectively discards tokens with low confidence scores, such as low log-probability or high entropy, relative to others in the current iteration. This process is repeated until all tokens are finalized, allowing for a more adaptive number of iterations. Some approaches further combine low-confidence filtering with block-wise remasking to support in-block autoregressive generation.

In contrast, our proposed $S^3$ method adopts a simple yet effective top-$K$ remask strategy, where $K = O/n$, with $O$ denoting the total number of tokens to be generated and $n$ being a tunable number of denoising steps. Compared to the block-wise remasking method, our approach is significantly more efficient and introduces an additional level of controllability by predefining the number of generation iterations, offering a practical trade-off between speed and output quality.

\section{Prompts}

This section provides the complete prompts used in our experiments for both the baseline method and our proposed Self-adaptive Schema Scaffolding ($S^3$) approach.

The baseline method employs a detailed prompt that explicitly specifies the desired output structure through a comprehensive JSON schema, as shown in Figure~\ref{fig:baseline_prompt}. This approach requires the model to understand and adhere to the predefined schema constraints based solely on the textual instructions.

In contrast, our Self-adaptive Schema Scaffolding ($S^3$) method utilizes a significantly simplified prompt (Figure~\ref{fig:s^3_prompt}) that omits explicit structural specifications. Instead, our method relies on the inherent structural constraints enforced by the scaffolding mechanism, demonstrating the token efficiency and clarity of $S^3$.

\begin{figure}[!ht]
\centering

\begin{tcolorbox}[width=\linewidth, fonttitle = \small\bfseries, title=Baseline Prompt ,colframe=gray!2!black,colback=gray!2!white,boxrule=1pt,boxsep=0pt,left=5pt,right=5pt,fontupper=\footnotesize, halign title = flush center]
\textbf{Instruction:} 

Extract the following information from the provided document and return only a JSON response with no additional text or explanation: name, birth\_date, birth\_place, death\_date, death\_place, nationality, occupation.

The response must conform to this JSON schema:

\begin{lstlisting}[basicstyle=\tiny\ttfamily, breaklines=true, frame=single, frameround=tttt]
{
  "schema": "http://json-schema.org/draft-07/schema\#",
  "type": "object",
  "properties": {
    "name": {
      "type": ["string", "null"],
      "description": "Full name of the person"
    },
    "birth\_date": {
      "type": ["string", "null"],
      "description": "Birth date in ISO format (YYYY-MM-DD, YYYY-MM, or YYYY)"
    },
    "birth\_place": {
      "type": ["string", "null"],
      "description": "Place of birth (city, country format preferred)"
    },
    "death\_date": {
      "type": ["string", "null"],
      "description": "Death date in ISO format (YYYY-MM-DD, YYYY-MM, or YYYY)"
    },
    "death\_place": {
      "type": ["string", "null"],
      "description": "Place of death (city, country format preferred)"
    },
    "nationality": {
      "type": ["string", "null"],
      "description": "Nationality or citizenship"
    },
    "occupation": {
      "type": ["string", "null"],
      "description": "Primary occupation or profession"
    }
  },
  "required": ["name", "birth\_date", "birth\_place", "death\_date", "death\_place", "nationality", "occupation"]
}
\end{lstlisting}

If any information is not available in the document, use null for that field.

[\textit{DOCUMENT TEXT}]
\vspace{0.5em}


\end{tcolorbox}

\caption{The complete prompt we use for the baseline method.}
\label{fig:baseline_prompt}
\end{figure}

\begin{figure}[tb]
\centering

\begin{tcolorbox}[width=\linewidth, fonttitle = \small\bfseries, title=$S^3$ Prompt ,colframe=gray!2!black,colback=gray!2!white,boxrule=1pt,boxsep=0pt,left=5pt,right=5pt,fontupper=\footnotesize, halign title = flush center]
\textbf{Instruction:} 

Extract information from the provided document and return only a JSON response with no additional text or explanation. If any information is not available in the document, use \texttt{null} for that field. [\textit{DOCUMENT TEXT}]
\vspace{0.5em}

\end{tcolorbox}

\caption{The complete prompt for our Self-adaptive Schema Scaffolding ($S^3$) method. Unlike the baseline approach, our method does not require explicit structural information in the prompt, as it naturally enforces predefined structure constraints on the model's output.}
\label{fig:s^3_prompt}
\end{figure}

\section{Proof}
\subsection{Proof of Theorem.~\ref{thm:scaffold_convergence}}
\label{proof:scaffold_convergence}
\begin{proof}
The diffusion model learns to predict $p_{\phi}(x_0^i|x_t)$ for each masked position $i$ where $x_t^i = \mathbf{M}$. With structural scaffolding, we partition positions into:
- Fixed scaffold positions: $\mathcal{S} = \{i : x_t^i \neq \mathbf{M}, \text{ fixed by structure}\}$
- Variable positions: $\mathcal{V} = \{i : x_t^i = \mathbf{M}\}$

The model's prediction at each masked position depends on the context:
$$p_{\phi}(x_0^i|x_t) = p_{\phi}(x_0^i|\{x_t^j\}_{j \neq i})$$

With scaffolding, the context includes correct structural tokens at positions $\mathcal{S}$, providing stronger signal:
$$p_{\phi}(x_0^i|x_t \text{ with } \mathcal{S}) = p_{\phi}(x_0^i|\mathcal{S} \cup \{x_t^j\}_{j \in \mathcal{V} \setminus \{i\}})$$

Since the scaffold tokens are correct by construction (they match the target structure), they reduce uncertainty in the conditional distribution. The error reduction is proportional to the informativeness of the scaffold.

For each masked position, the expected error with scaffolding is:
$$\mathbb{E}[\epsilon_i | \mathcal{S}] \leq \mathbb{E}[\epsilon_i] \cdot (1 - I(\mathcal{S}; x_0^i))$$

where $I(\mathcal{S}; x_0^i)$ is the mutual information between scaffold and target token.

Aggregating over all masked positions and noting that structural tokens typically have high mutual information with content tokens (e.g., field names predict value types), we get:
$$\mathbb{E}[\|\hat{\mathbf{x}}_0 - \mathbf{x}_0\|_{\mathcal{M}}] \leq \mathbb{E}[\|\tilde{\mathbf{x}}_0 - \mathbf{x}_0\|_{\mathcal{M}}] \cdot \left(1 - \frac{|\mathcal{S}|}{L}\right)$$

The equality holds when scaffold and content are independent, which rarely occurs in structured generation.
\end{proof}



\end{document}